\documentclass[10pt,twocolumn,letterpaper]{article}

\usepackage{cvpr}
\usepackage{times}
\usepackage{epsfig}
\usepackage{graphicx}
\usepackage{subfigure}
\usepackage{multirow}
\usepackage{amsmath}
\usepackage{amssymb}

\usepackage[breaklinks=true,bookmarks=false]{hyperref}

\cvprfinalcopy 

\def\cvprPaperID{****} 

\begin{document}

\title{Integration of Text-maps in Convolutional Neural Networks \\ for Region Detection among Different Textual Categories}

\author{Roberto Arroyo, Javier Tovar, Francisco J. Delgado, \\ Emilio J. Almaz\'an,
Diego G. Serrador, Antonio Hurtado\\
Nielsen Connect AI\\
Calle Salvador de Madariaga, 1, 28027, Madrid, Spain\\
{\tt\small Corresponding author: roberto.arroyo@nielsen.com}}

\maketitle

\begin{abstract}
In this work, we propose a new technique that combines appearance and text in a Convolutional Neural Network~(CNN), with the aim of detecting regions of different textual categories. We define a novel visual representation of the semantic meaning of text that allows a seamless integration in a standard CNN architecture. This representation, referred to as \mbox{text-map}, is integrated with the actual image to provide a much richer input to the network. \mbox{Text-maps} are colored with different intensities depending on the relevance of the words recognized over the image. Concretely, these words are previously extracted using Optical Character Recognition (OCR) and they are colored according to the probability of belonging to a textual category of interest. In this sense, this solution is especially relevant in the context of item coding for supermarket products, where different types of textual categories must be identified, such as ingredients or nutritional facts. We evaluated our solution in the proprietary item coding dataset of Nielsen Brandbank, which contains more than 10,000 images for train and 2,000 images for test. The reported results demonstrate that our approach focused on visual and textual data outperforms \mbox{state-of-the-art} algorithms only based on appearance, such as standard \mbox{Faster R-CNN}. These enhancements are reflected in precision and recall, which are improved in 42 and 33 points respectively.
\end{abstract}

\section{Introduction}

The increasing popularity of deep learning has revolutionized several computer vision areas. In this regard, the detection of regions of interest over images has greatly improved since the rise of methods based on Convolutional Neural Networks (CNNs), such as \mbox{Faster R-CNN}~\cite{ref:Ren15nips}. However, these techniques rely on visual cues to discriminate regions and they are not expected to achieve high performance when the appearance of the regions is not discriminative. A typical example of this is the detection of regions with different categories of text, where the most distinguishing information resides in the semantic meaning of the text.

\begin{figure} \includegraphics[width=\columnwidth]{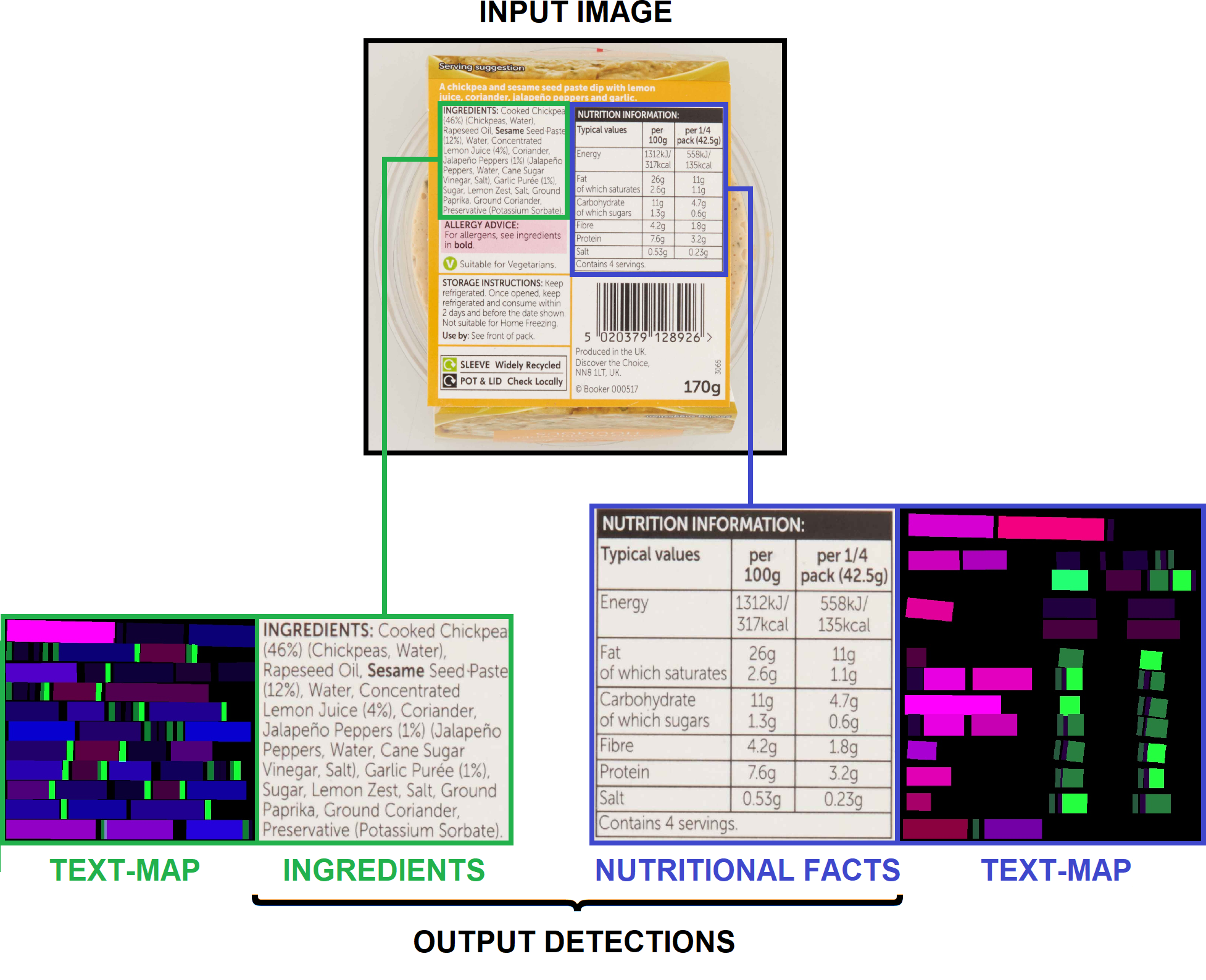} \caption{A brief overview about item coding based on ingredients and nutritional facts. The diagram depicts a processed image from a product packaging jointly with an introduction to the visual representation of the computed \mbox{text-maps} for each region of interest.} 
\label{fig:Overview} 
\end{figure}

Within this context, automated item coding is a specific case in which appearance is not enough. Item coding refers to the process of transcribing the characteristics of an item into a database. For instance, a common application of item coding is the extraction of data from supermarket products. As shown in Figure~\ref{fig:Overview}, different types of textual information are commonly detected, such as ingredients or nutritional facts. In this example, we can see how the textual information of the regions plays a differential role. Then, the fusion of textual representations with \mbox{appearance-based} techniques is a promising alternative with respect to traditional computer vision approaches for item coding~\cite{ref:Kulyukin13ipcv,ref:Venugopal18aicc}.

In the state of the art of image recognition, it is common to see works that combine data from different sources, where each one provides complementary information to create a richer input data. For instance, some examples are related to the combination of RGB with depth maps~\cite{ref:Arroyo14iros,ref:Eitel15iros}. Besides, other approaches build upon the fusion of RGB with raw text data in varied use cases~\cite{ref:Bai18ieeea,ref:Prasad18eccv}.

In this paper, we present a novel method that transforms textual information into an \mbox{image-based} representation named \mbox{text-map}, as visually introduced in Figure~\ref{fig:Overview}. This concept allows a seamless integration in standard CNNs. In this sense, our approach does not modify the internal architecture because it only extends the inputs of the network, which is an advantage for adaptability to different problems. In Section~\ref{sec:cnn_method}, we describe in detail our CNN proposal, jointly with the associated concept of \mbox{text-map}. In Section~\ref{sec:experiments}, we present several experiments and results for item coding to validate our approach based on \mbox{text-maps} with respect to methods only based on appearance, such as standard~\mbox{Faster R-CNN}. Finally, we highlight the main conclusions derived from this work and some future research lines in Section~\ref{sec:conclusions}.

\section{Our Method for Combining Appearance and Textual Information in a CNN} \label{sec:cnn_method}

The key of our approach for region detection among different textual categories resides in the generation of \mbox{text-maps}. Then, it is essential to describe their main characteristics and how they are inputted into a CNN jointly with the RGB image information.

\subsection{Generation of Text-maps for Injecting Textual Information}

\begin{figure}[!ht] 
\centering
\subfigure[Use case of ingredients.]                                     
{\includegraphics[width=\columnwidth]{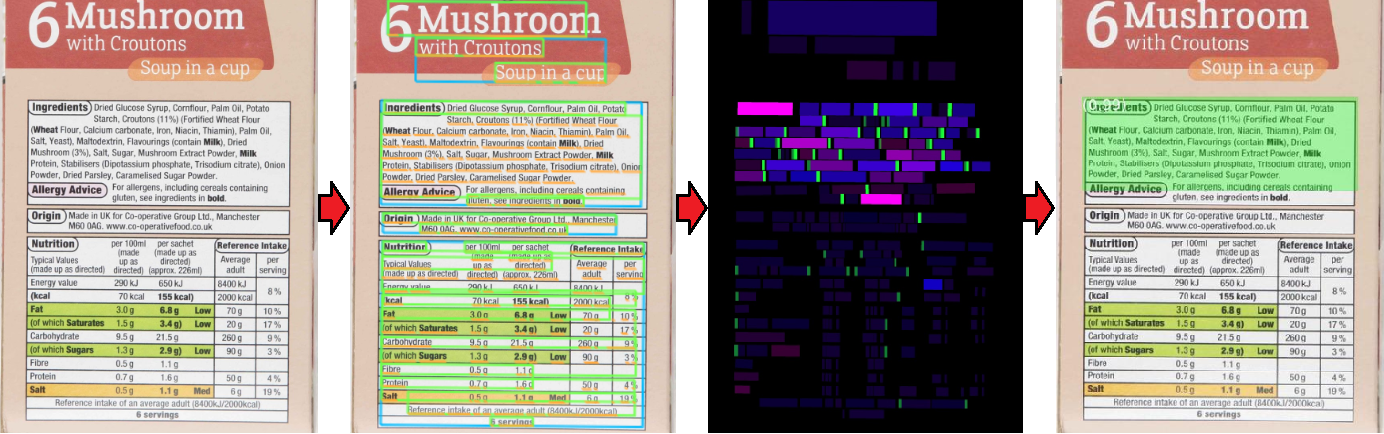}}
\subfigure[Use case of nutritional facts.]                                     
{\includegraphics[width=\columnwidth]{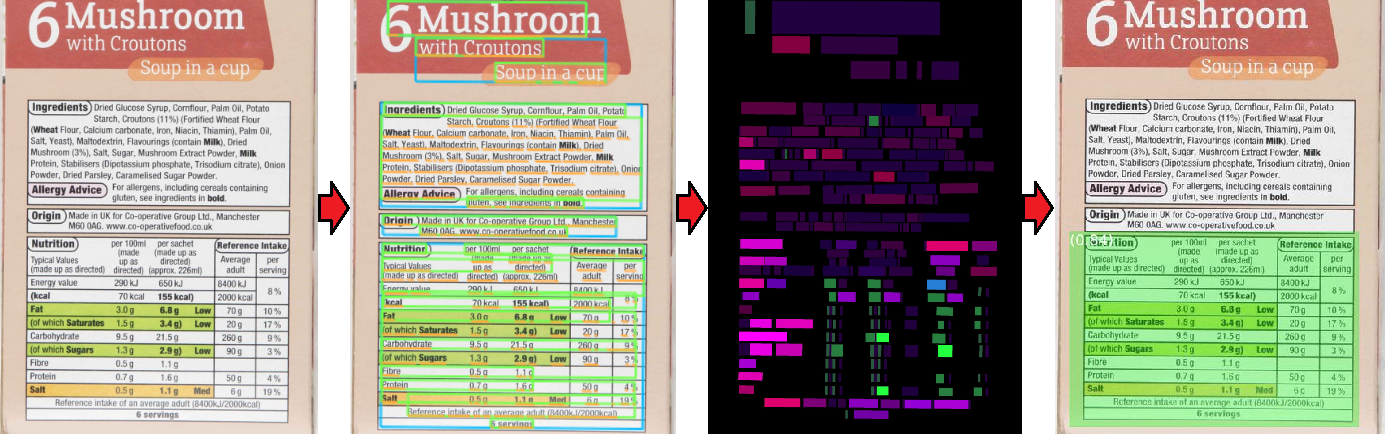}}
\caption{Stages for obtaining the text-maps required for detecting the regions of interest. 1)~Input images. 2)~OCR~extraction. 3)~Text-maps~extraction. 4)~Output detection.} 
\label{fig:Text-maps_stage} 
\end{figure}

We define a \mbox{text-map} as a visual representation of the original image where the zones that contain words are colored with different intensities depending on the relevance of the word. More specifically, the algorithm colors the text zone retrieved from a standard OCR engine (e.g.,~Google~Cloud~Vision~API~\footnote{\url{https://cloud.google.com/vision/docs/ocr}}) according to the probability of the text to belong to a certain category of interest. In our item coding system, these categories of interest are ingredients and nutritional facts, as depicted in the examples presented in Figure~\ref{fig:Text-maps_stage}. For instance, a zone that contains the word \emph{milk} will have a high probability of belonging to ingredients and will be brightly colored in its respective \mbox{text-map}. Similarly, the word \emph{protein} will be brightly colored in the \mbox{text-map} corresponding to nutritional facts.

In the specific use case of our item coding approach, we use \mbox{text-maps} composed of 3 channels. Each channel encodes the relevance of each word based on different metrics, which are computed as follows:

\begin{itemize}
\item \emph{Red channel}: Occurrences of a specific word inside the \mbox{ground-truth} regions vs total occurrences.

\item \emph{Green channel}: Typical punctuation signs in the regions of interest. Commas or parentheses are considered for ingredients. Numerical values and related symbols (e.g., '$\%$') are used for nutritional facts.

\item \emph{Blue channel}: Bayesian distance between text and dictionaries of ingredients and nutrients. These dictionaries are generated using \mbox{ground-truth} data.
\end{itemize}

\subsection{Design of the Proposed CNN Approach}

\begin{figure} \includegraphics[width=\columnwidth]{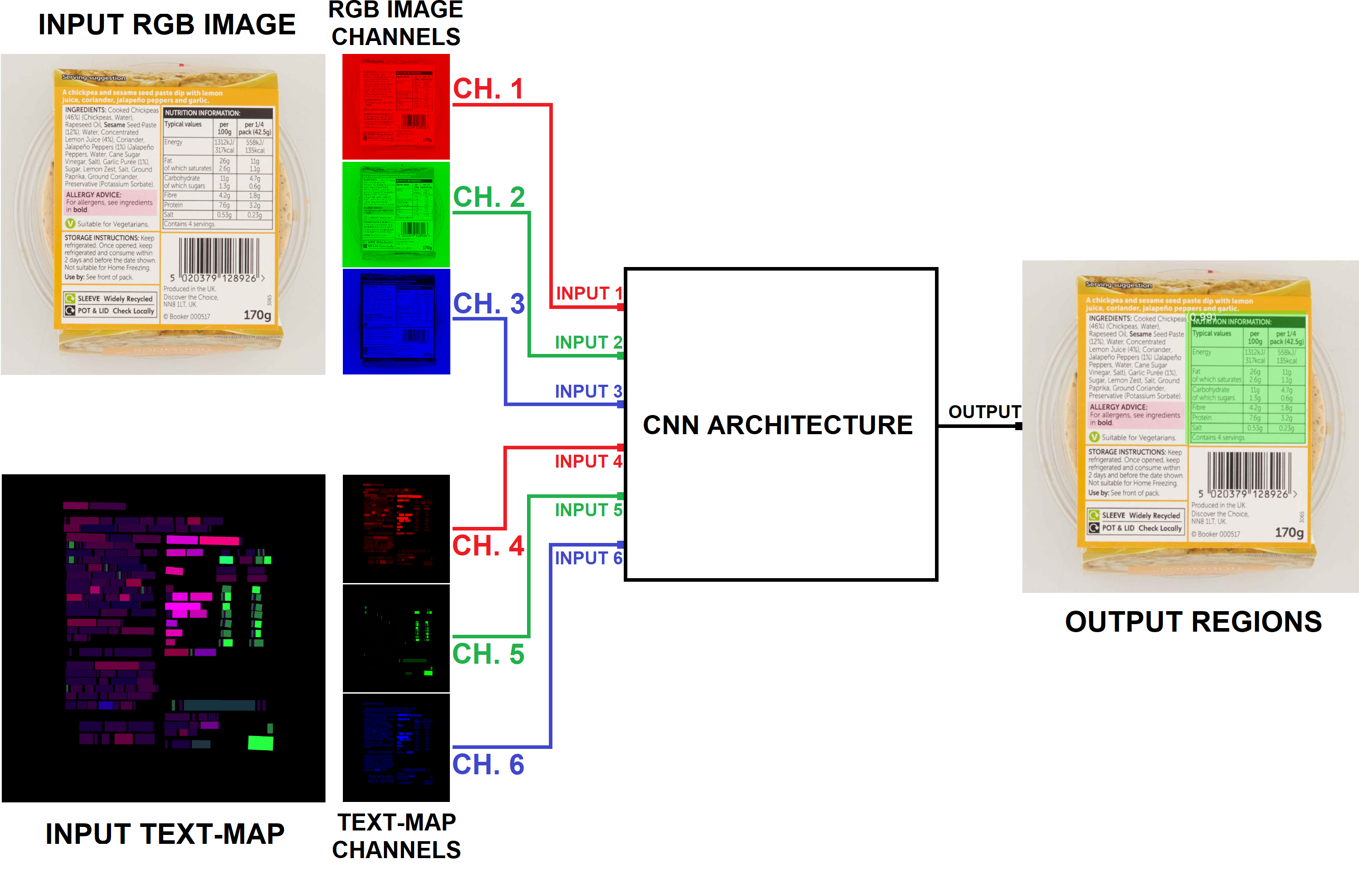}
\caption{Proposed CNN with RGB channels plus \mbox{text-map} channels. The presented example is focused on nutritional facts, but it is analog for ingredients. In our use case, we use 3 \mbox{text-map} channels as input, but this number could be changed depending on the problem requirements.} 
\label{fig:CNN_architecture} 
\end{figure}

After generating a \mbox{text-map}, it is injected into the CNN architecture jointly with the original image to detect its specific text regions of interest. Standard CNN architectures receive 3 RGB channels as input. In this regard, the architecture proposed in our item coding system receives the 3 RGB channels plus 3 channels for the \mbox{text-maps}, so it applies 6 channels in total, as represented in Figure~\ref{fig:CNN_architecture} for the nutritional facts case (analog for ingredients).

The core of our CNN model can use any standard backbone network. In the case of our item coding approach, ResNet~\cite{ref:He16cvpr} exhibited a satisfactory performance as CNN backbone for the applied architecture. Obviously, the CNN model must be trained using previously labeled data for a proper performance. In this sense, the training of our models is started from \mbox{pre-trained} weights  obtained from ImageNet, which is a robust dataset commonly applied in some of the most representative proposals in the state of the art~\cite{ref:Krizhevsky12nips}. Besides, several works have demonstrated the great transferability of CNN features from different datasets and problems, as studied in~\cite{ref:Yosinski14nips}.

In addition, it must be noted that after predicting the text regions of interest using the described CNN proposal, our system easily obtains the resultant ingredients and nutritional facts by \mbox{post-processing} the OCR previously computed for \mbox{text-maps} within these predicted regions. However, it is out of scope of this paper and we prefer to focus on the region detection among different text categories, which is our main contribution presented in the topic of language and vision fusion.

\section{Experiments and Results} \label{sec:experiments}

With the aim of validating the performance of our CNN model with respect to \mbox{state-of-the-art} approaches only based on visual appearance, we present the main experiments carried out for testing our method based on combined visual and textual information.

\subsection{Dataset for Item Coding}

The acquisition of images and labeled data for evaluating our automated item coding system is a costly process. Due to this, there are not large datasets publicly available for these tasks. We found recent public datasets focused on the recognition of merchandise in supermarkets, such as the D2S dataset~\cite{ref:Follmann18eccv}. Unfortunately, this dataset is not suitable for our item coding tests because of the long distance between camera and products, so ingredients and nutritional facts can not be visually distinguished and labeling for them is not available. Then, we have used our own labeled data from Nielsen 
Brandbank\footnote{\url{https://www.brandbank.com/}} to train and evaluate our automated item coding solution. This dataset is composed of more than 10,000 labeled images for training and 2,000 images for validation and test.

\subsection{Training and Hyperparameters}

To train our CNN model, we set up an architecture based on \mbox{Faster R-CNN} with \mbox{ResNet-101} as backbone using as input data the combination of image and \mbox{text-map} channels. The following main hyperparameters are applied during $10$ epochs: learning~rate~=~$1\cdot10^{-5}$, weight~decay~=~$1\cdot10^{-6}$, dropout keep prob~=~$0.8$, batch~size~=~$1$ and Adam~optimizer~\cite{ref:Kingma15iclr}. We compare our solution against a model trained with a standard \mbox{Faster R-CNN} only based on appearance. Analog hyperparameters are also used in this case to perform a fair comparison.

\subsection{Quantitative Results}

Our experiments are mainly focused on precision and recall results for ingredients and nutritional facts detection. In Table~\ref{table:results}, it can be seen how our CNN model based on \mbox{text-maps} clearly outperforms standard \mbox{Faster R-CNN}. Concretely, our approach increases precision and recall in 42 and 33 points respectively. Besides, we enhance in 38 points the total accuracy, which is calculated as the division of true positives between the sum of true positives, false positives and false negatives. According to these results, the improvements given by our solution are demonstrated for region detection among different textual categories.
\begin{table*}
\centering
\begin{tabular}{l|c|c|c|c|c|c|}
\cline{2-7}
\multirow{2}{*}{}                                & \multicolumn{3}{c|}{\textbf{\mbox{Faster R-CNN} (3 channels)}}   & \multicolumn{3}{c|}{\textbf{Ours (6 channels)}}          \\ \cline{2-7} 
                                                 & \textbf{Precision} & \textbf{Recall} & \textbf{Accuracy} & \textbf{Precision} & \textbf{Recall} & \textbf{Accuracy} \\ \hline
\multicolumn{1}{|l|}{\textbf{Ingredients}}       & 0.25               & 0.31            & 0.15              & \textbf{0.70}      & \textbf{0.73}   & \textbf{0.56}     \\ \hline
\multicolumn{1}{|l|}{\textbf{Nutritional facts}} & 0.34               & 0.57            & 0.27              & \textbf{0.72}      & \textbf{0.81}   & \textbf{0.62}     \\ \hline
\multicolumn{1}{|l|}{\textbf{Totals}}            & 0.29               & 0.44            & 0.21              & \textbf{0.71}      & \textbf{0.77}   & \textbf{0.59}     \\ \hline
\end{tabular}
\vspace{2mm}
\caption{Standard \mbox{Faster R-CNN} vs our model with text-maps. A confidence threshold of $0.7$ is applied in both cases for processing results.}\label{table:results}
\end{table*}

\subsection{Qualitative Results}

\begin{figure*}
\centering
\subfigure[Detections using standard \mbox{Faster R-CNN}.]                                     
{\includegraphics[width=\textwidth]{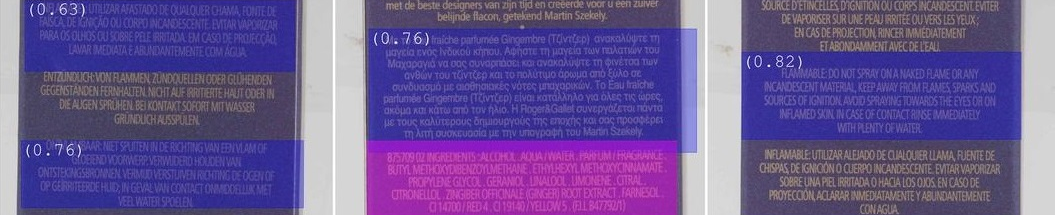}}
\subfigure[Detections using our CNN model based on visual and textual information.]                                     
{\includegraphics[width=\textwidth]{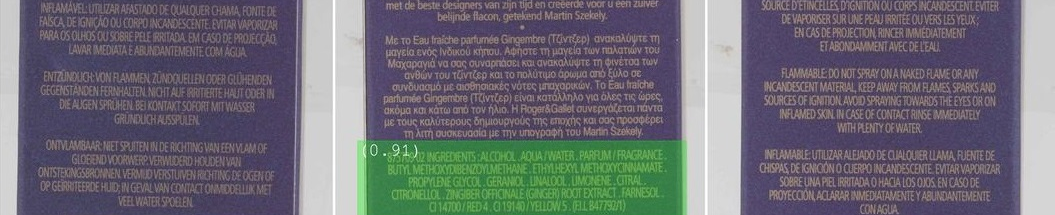}}
\caption{A visual example of the detection of ingredients regions using a model trained with a standard \mbox{Faster R-CNN} vs our approach. False negatives are marked in fuchsia, false positives in blue and true positives in green. Confidences are depicted in the bounding boxes.} 
\label{fig:Example_detection} 
\end{figure*}

Apart from quantitative results, we also depict some qualitative visual results in Figure~\ref{fig:Example_detection}. In this example, a lot of false positives are wrongly detected by the standard \mbox{Faster R-CNN} model due to the similarity of the visual appearance in the different types of text. However, our CNN model based on visual and textual information is able to 
correctly predict the ingredients region and false positives are not detected. These results evidence again the suitability of our CNN solution for item coding problems.

\section{Conclusions and Future Works} \label{sec:conclusions}

The application of language as a mechanism to structure and reason about visual perception has been demonstrated along this paper. In this regard, our innovative CNN model enriched with \mbox{text-maps} has evidenced its effectiveness with respect to \mbox{state-of-the-art} solutions only based on visual appearance. We presented results related to our specific use case for item coding, but the concept of \mbox{text-maps} is applicable to other problems focused on the detection of regions with different textual categories.

In further works, other text regions of interest for item coding are planned to be detected by our system, such as storage information or cooking instructions, among others. Moreover, the proposed technique for generating \mbox{text-maps} and the number of channels could be adapted to other text region detection challenges in future researches.

{\small
\bibliographystyle{ieee_fullname}
\bibliography{bibliography}
}

\end{document}